\documentclass{article}

\usepackage{PRIMEarxiv}

\usepackage[utf8]{inputenc} 
\usepackage[T1]{fontenc}    
\usepackage{hyperref}       
\usepackage{url}            
\usepackage{booktabs}       
\usepackage{amsfonts}       
\usepackage{nicefrac}       
\usepackage{microtype}      
\usepackage{lipsum}
\usepackage{fancyhdr}       
\usepackage{graphicx}       
\graphicspath{{media/}}     
\usepackage{amsmath}

\usepackage{graphicx}
\usepackage{amssymb}
\usepackage{hyperref}  
\usepackage[acronym]{glossaries}
\usepackage{glossary-inline}

\usepackage{subcaption}
\usepackage{lscape} 

\usepackage{algorithmic}
\usepackage{algorithm}
\usepackage{booktabs}
\usepackage{multirow}
\usepackage[official]{eurosym}
\usepackage{makecell}
\usepackage{subcaption}
\usepackage{xcolor}
\usepackage[labelfont=bf]{caption}
\usepackage{setspace}
\usepackage{stfloats}
\usepackage{bigfoot}
\usepackage{multirow}

\makeglossaries
\newacronym{drl}{DRL}{Deep Reinforcement Learning}
\newacronym{rl}{RL}{Reinforcement Learning}
\newacronym{ppo}{PPO}{Proximal Policy Optimization}
\newacronym{re}{RE}{Renewable Energy}
\newacronym{ml}{ML}{Machine Learning}
\newacronym{ca}{CAgent}{CurriculumAgent}
\newacronym{pbt}{PBT}{Population Based Training}
\newacronym{mdp}{MDP}{Markov Decision Process}
\newacronym{sgd}{SGD}{Stochastic Gradient Decent}
\newacronym{tso}{TSO}{Transmission System Operators}
\newacronym{opf}{OPF}{Optimal Power Flow}
\newacronym{l2rpn}{L2RPN}{Learning to Run a Power Network}
\newacronym{gnn}{GNN}{Graph Neural Network}
\newacronym{mst}{MST}{Median Survival Time}
\newacronym{tts}{TTs}{Target Topologies}
\newacronym{tt}{TT}{Target Topology}
\newacronym{mstcm}{MSTCM}{Median Survival Time across Chronic Medians}
\newacronym{sacd}{SACD}{Soft Actor-Critic Discrete }
\newacronym{senior}{$Senior_{95\%}$}{Senior Agent}
\newacronym{topo}{$TopoAgent_{85-95\%}$}{Topology Agent} %
\newacronym{senior85}{$Senior_{85\%}$}{85\%-Senior Agent}
\newacronym{d_n}{$DoNothing$}{Do-Nothing Agent}
\newacronym{smaac}{SMAAC}{Semi Markov Afterstate Actor Critic}
\newacronym{ddqn}{DDQN}{Dueling Deep Q Network}
\newacronym{bohb}{BOHB}{Bayesian Optimization Hyperband}

\title{HUGO - Highlighting Unseen Grid Options: Combining Deep Reinforcement Learning with a Heuristic Target Topology Approach
}
\author{
  Malte Lehna \thanks{\textbf{Corresponding Author}: \texttt{malte.lehna@iee.fraunhofer.de}, \url{https://orcid.org/0000-0003-0621-1442}} \\
  Fraunhofer IEE, Kassel University \\
  Kassel \\
  Germany\\
  \\
   \And
  Clara Holzhüter\thanks{ \url{https://orcid.org/0000-0001-8365-5544}} \\
  Fraunhofer IEE, Kassel University \\
  Kassel \\
  Germany\\
  \And
  Sven Tomforde\thanks{\url{https://orcid.org/0000-0002-5825-8915}} \\
  Kiel University: Intelligent Systems \\
  Kiel \\
  Germany\\
    \And
  Christoph Scholz\thanks{\url{https://orcid.org/0000-0002-8719-8261}} \\
  Fraunhofer IEE, Kassel University \\
  Kassel \\
  Germany\\
}

\begin{document}
\maketitle

\begin{abstract}
With the growth of \gls{re} generation, the operation of power grids has become increasingly complex. One solution could be automated grid operation, where \gls{drl} has repeatedly shown significant potential in \gls{l2rpn} challenges.
However, only individual actions at the substation level have been subjected to topology optimization by most existing \gls{drl} algorithms. 
In contrast, we propose a more holistic approach by proposing specific \gls{tts} as actions. These topologies are selected based on their robustness. As part of this paper, we present a search algorithm to find the \gls{tts} and upgrade our previously developed \gls{drl} agent \gls{ca} to a novel topology agent. We compare the upgrade to the previous \gls{ca} and can increase their \gls{l2rpn} score significantly by 10\%. Further, we achieve a 25\% better median survival time with our \gls{tts} included. Later analysis shows that almost all \gls{tts} are close to the base topology, explaining their robustness.
\end{abstract}


\section{Introduction}
\label{sec:intro}
As the fight against climate change intensifies, power generation has changed dramatically in recent years. Increasing amounts of \gls{re} is produced, resulting in more variable injections to the grid. In addition, the overall power generation is becoming more decentralized. 
This poses new challenges for the \gls{tso} as they have to become flexible with their current electricity grid to handle the volatility. Previously, redispatch and curtailment has been the method of choice for achieving grid stability. However, this approach is costly and often induces further CO2 emissions. A relatively new approach in the research community is now bus switching at the substation level to change the topology of the grid, which is a more cost-effective alternative \cite{marot2020learning}. As mentioned in \cite{bacher1986network}, intelligent switching in critical areas of the grid allows overloads to be diverted to stabilize the grid to some extent. 
To solve this problem, researchers propose applying deep learning methods, particularly in \gls{drl}, which could significantly reduce computational costs of the grid optimization \cite{viebahn2022potential}. Such approaches were first tested in the \glsfirst{l2rpn} challenge \cite{marot2020learning} by the French \gls{tso} RTE.\footnote{\gls{l2rpn} challenges: \url{https://l2rpn.chalearn.org/} (last accessed 05/04/2024).} 
Through their continuous development, the challenges, and the realistic representation of electricity grids, \gls{l2rpn} has become the leading benchmark for \gls{drl} based grid simulations in the community and is used by various researchers\cite{marot2020learning,marot2021learning}.

\subsection{Main Contributions}
\label{ssec:contribution}
\begin{figure*}[b]
\setstretch{0.95}
    \centering
    \caption{Simplified example of a topology action based on \cite{marot2020learning}. With the injections of the generators and the high demand of the loads, the grid has an overload in the right line (red) in time step $t$ (Box \textbf{a}). To reach the stable state in $t+1$, splitting the bottom-right substation by assigning two different buses is essential (Box \textbf{b}). This can be achieved by executing the substation action (Box \textbf{c}) on substation No. 4. Alternatively, one can describe the desired outcome in the form of a \gls{tt} (Box \textbf{d}). Note that the blue dots represent the $bus_{one}$, the red ones $bus_{two}$ and the dotted lines  representation separate substations. }
    \includegraphics[trim={0cm 12cm 0.0cm 0.0cm},clip, width=1.0\linewidth]{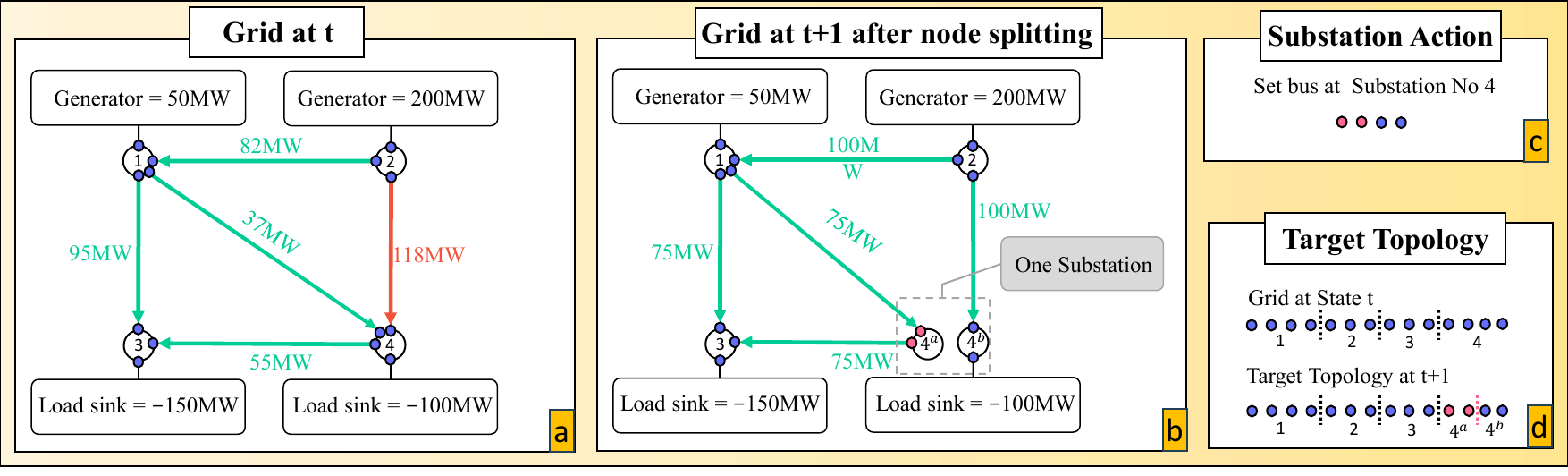}
    \label{fig:topo_act}
\end{figure*}

Previously,\gls{drl} researchers often considered actions that change bus configurations at the substation level to change the grid topology. 
These actions affect only one substation but can affect multiple buses on that substation. We refer to these actions as substation actions. The problem is that these actions are often considered isolated for only the current time-step. While they may be beneficial for the current stage, they may lead to unstable or suboptimal topologies in the long run. In reality, \gls{tso} operators do not look at independent substation actions, but instead consider changing a combination of substations over multiple time steps. This holistic approach is necessary to reach the multiple objectives of the \gls{tso}, as described in \cite{viebahn2024gridoptions}. However, these holistic topology approaches are rarely addressed in \gls{drl} research for grid optimization. This may be caused by the design of the leading framework Grid2Op \cite{grid2op}.
\\
In our work, we propose a different approach. Instead of considering actions that switch only one single substation, we are looking at the overall topology of the electricity grid, i.e., the current configuration of all buses at all substations. The idea is that there are certain \gls{tts} that are more robust than other topologies. This is inspired by the findings of \cite{viebahn2024gridoptions}, who derive the necessity of \gls{tts} directly from the \gls{tso} objectives. If our current topology is not resilient enough,  we try to reach a close \gls{tts} instead. Since we can reach the \gls{tt} from almost any topology configuration, we do not need to learn specific combinations of substation actions. This is especially helpful in more complex grids, as the \gls{tts} may lead to sequential execution of multiple substation actions. In Figure \ref{fig:topo_act}, we visualize the different approaches of the substation action and the \gls{tt} using a simple example.
As a contribution of this work, we propose a search algorithm to find adequate \gls{tts}. With the algorithm, we identify \gls{tts} that prove to be robust against instability, given an existing set of substation actions. 
In addition, we extend our previously published \glsfirst{ca}
approach \cite{lehna2023managing} to a Topology Agent by adding a greedy search component with \gls{tts}. We test our agent on the validation grid of WCCI 2022 \gls{l2rpn} challenge to ensure that our approach is beneficial for topology optimization.

The contributions can be summarized as follows: \glsreset{tts} 
\begin{itemize}
    \item We provide the retrained \gls{ca}\cite{lehna2023managing} on the WCCI 2022 environment with a substation action set of 2030 actions. 
    \item In light of the concept of \gls{tts}, we present our novel search algorithm to find \gls{tts} based on existing substation actions.
    \item We improve our previous \gls{ca} to a \gls{topo} agent by adding a greedy component that iterates through \gls{tts} and selects the best \gls{tt}.
    \item We show a significant increase by more than 10\% in the mean \gls{l2rpn} score performance.
    \item In addition, we observe that the \gls{topo} has a 25\% higher median survival time than the previous \gls{ca}. 
    \item All extensions will be published upon acceptance as part of the \gls{ca} Github repository.
\end{itemize}

The rest of the paper is structured as follows. First, we present the related work in Section \ref{sec:rel_work}, followed by a description of the environment and the \gls{ca} in Section \ref{sec:env_agent}. We then propose our new \gls{tt} approach in detail, with the topology search and the topology agent in Section \ref{sec:topo}. We evaluate the topology agent on the benchmark scenarios in Section \ref{sec:experiment}, followed by a discussion of the results in Section \ref{sec:discussion}. Finally, we conclude in Section \ref{sec:conclusion}.

\section{Related Work}
\label{sec:rel_work}
After the pioneering breakthrough of \gls{drl} in various Atari games in 2015 \cite{mnih2015human,mnih2016asynchronous}, \gls{drl} has demonstrated exceptional performance in several research areas \cite{lee2020learning,schrittwieser2020mastering,berner2019dota}.
As a result, there has been a growing interest in adopting \gls{drl} approaches in the field of grid control \cite{viebahn2022potential}. This is particularly evident in the context of the \gls{l2rpn} challenge \cite{marot2020l2rpn,kelly2020reinforcement,9494879,marot2021learning}, which had a wide response in the power system and \gls{ml} community. With their Grid2Op environment\cite{grid2op} they introduced a realistic benchmark, which allows researchers to validate their grid control approaches on multiple simulation environments following the GYM framework of OpenAI\cite{brockman2016openai}. While there have been some purely rule-based agents, e.g. \cite{marot2018expert}, the challenges have primarily been won by combining \gls{drl} with rule-based elements. 
In this regard, the first effective \gls{drl} approach to grid control has been introduced by the winners of the 2019 challenge \cite{lan2020ai}. Their approach was based on a \gls{ddqn} and combined an imitation learning strategy with guided exploration. 
Following a more complex \gls{l2rpn} challenge of 2020 \cite{marot2020l2rpn}, \cite{zhou2021action} proposed a \gls{drl} approach that combines a planning algorithm with an evolutionary strategy. The available actions provided by the policy were searched by the planning algorithm, which was then optimized using the evolutionary algorithm. Gaussian white noise was added to ensure adequate exploration of the policies\cite{zhou2021action}. The second best performance of the robustness track has been achieved by\cite{binbinchen}, which is the basis of the curriculum agent and further described in Section \ref{ssec:curriculumagent}. A similar approach also combining a \gls{drl} algorithm with a heuristic was presented by \cite{chauhan2022powrl}. Another approach worth mentioning was proposed by \cite{dorfer2022power}, who introduced a Monte Carlo tree search approach to finding appropriate actions, similar to the AlphaZero algorithm in \cite{silver2017mastering}. With their agent, they were able to win the \gls{l2rpn} WCCI Challenge of 2022. Moreover, in their \cite{silver2017mastering} paper, they evaluated the effect of topology optimization with substation actions and showed that inclusion can reduce redispatching costs by up to 60\%. 
This shows the potential of substation actions and is the reason we chose their action set for this paper.
Another group of approaches is hierarchical agents such as the one presented by \cite{van2023multi}. The authors introduce a multi-agent \gls{drl} framework based on \gls{sacd} and \gls{ppo}. The approach optimizes power grid topologies through three levels. The highest level is rule-based that activates interventions only in critical conditions of the grid. The mid-level identifies and prioritizes actions for specific substations needing intervention. At the low level, substation-specific agents decide on bus configurations by applying \gls{drl} to make localized decisions independently. Two related approaches \cite{liu2024progressive, manczak2023hierarchical}also break down the decision process into three very similar stages. In particular, \cite{liu2024progressive} additionally used a \gls{gnn} to learn representations of the power system and its topology at different granularities before the multi-stage decision making. In contrast, \cite{hu2023towards} also followed a hierarchical approach but focused on fairness, i.e., distributing the long-term benefits among power plants fairly. They introduced a hierarchical optimization function and a reward system focusing on equitable supply-side gains.
Finally, we would like to highlight the work of the winners of the L2RPN WCCI 2020 challenge \cite{yoon2020winning}, who applied a \gls{smaac} algorithm with a \gls{gnn} to summarize their actions. 
Unlike other participants, they were among the first researchers to focus on whole topologies instead of isolated actions. Based on their initial findings, we take up their idea and propose a more holistic approach, including a detailed analysis of the effect of \gls{tts}.

\section{Environment and agent structure}
\label{sec:env_agent}
\subsection{Grid2Op Environment}
\label{ssec:grid2op}
\begin{figure*}
    \centering
    \caption{Visualization of the WCCI 2022 \gls{l2rpn} environment based on the IEEE118 grid. Grid2Op's internal plotting method was used to create the Figure.}
    \includegraphics[trim={0.0cm 0.5cm 0.0cm 0.5cm},clip, width=0.95\linewidth]{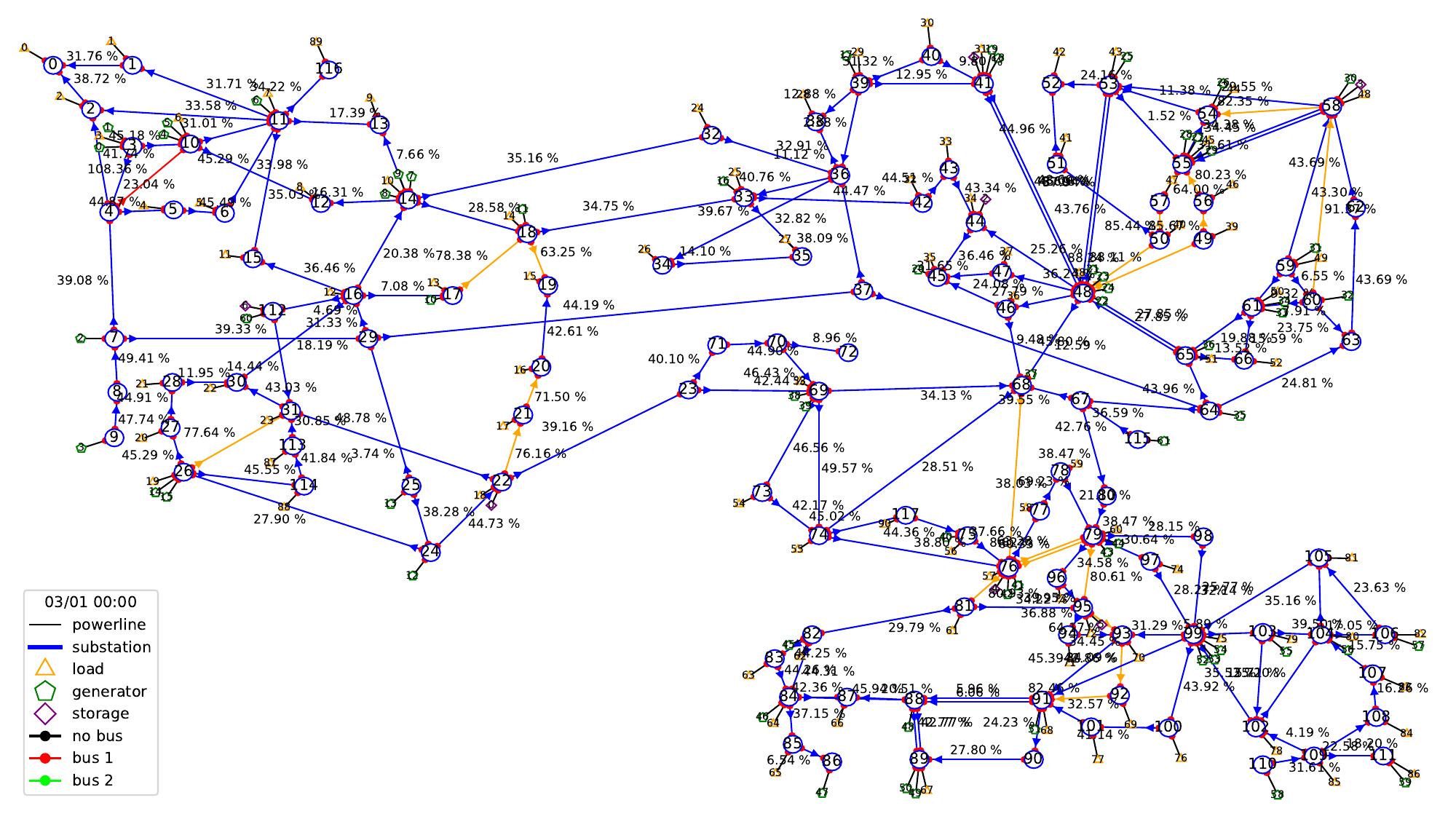}
    \label{fig:grid}
\end{figure*}
The Grid2Op package is the current state-of-the-art environment for \gls{drl} development related to electric grid operations and was developed by RTE France in the context of the \gls{l2rpn} challenges. 
Grid2Op includes multiple synthetic power grids, such as the IEEE14 or IEEE118 grids with a \verb|pandapower| \cite{pandapower.2018} or \verb|ligthsim2grid| \cite{lightsim2grid} backend, ensuring realistic load flow caluclations. To ensure that the environments follow a \gls{mdp}, they are modeled according to the OpenAI's Gym framework \cite{brockman2016openai} and are described in more detail in \cite{marot2020learning,kelly2020reinforcement,serre2022reinforcement}. 
In this paper, we use the \gls{l2rpn} WCCI 2022 environment, which is composed of the IEEE118 grid with an expected electricity mix of 2050, i.e., the share of fossil fuels for electricity generation is less than 3\% and the \gls{re} are drastically increased \cite{serre2022reinforcement}.
As seen in Figure \ref{fig:grid}, the grid has a total of 118 substations, 91 load sinks and 62 generators, as well as the seven battery storages, all connected by a total of 186 lines. As such, the observation space of the grid has the size $4295$ and contains various information about the power grid, such as active and reactive power flows, voltage magnitudes and voltage angles of the lines and substation buses. Moreover, the injections of the generators and storages, load demand, time variables, planed maintenance, cooldown periods and topology configurations are reported. The most important variable for this paper is the line capacity, which we denote as $\rho_{l,t} \in \mathbb{R}^+$ for line $l$ of all lines $l \in \mathcal{L}$ as well as the maximum capacity over all lines as $\rho_{max,t}=\max\limits_{l=1,\ldots,L} (\rho_{l,t})$. 
Besides observation, Grid2Op provides a simulation method with \verb|obs.simulate()| that can predict the state of the next step based on the current action, albeit with some margin of error. From this method we can derive the impact of the action on the maximum simulated line capacity $\hat{\rho}_{max,t+1}=\max\limits_{l=1,\ldots,L} (\hat{\rho}_{l,t+1})$, i.e., $\hat{\rho}_{l,t+1}$ is the simulated line capacity of $l$.
\newline
With respect to the action space, multiple action types are available, however, similar to \cite{lehna2023managing}, we are only interested in the substation actions, as we want to analyze the effect of topology changes on the robustness of the grid. The agents in this paper also use line actions to reconnect a disconnected line to the grid. Since they are simply implemented by a rule-based method, we do not discuss them further.
We describe the substation action as $a^{(bus)}$ from the set of all possible bus configurations on all substations $a^{(bus)} \in \ \mathcal{A}^{(bus)}$ with a total of $72957$ available substation actions. In Grid2Op, the elements of a substation can be connected to either $bus_{one}$ or $bus_{two}$, e.g, shown in Figure \ref{fig:topo_act}. By using different bus configurations, it is possible to control the power flow in the grid. However, the number of available bus permutations can become quite large with more elements at a substation and their optimization complex \cite{9494879}. 
Note that there are some restrictions on taking actions on previously modified lines and substations, as there are cooldown periods that limit interaction on the same element. There is also an adversarial agent \cite{omnes2021adversarial} that can disconnect lines quasi-randomly, causing additional cooldowns. 
\newline
There are two ways to terminate an episode in the environment. The first is a successful completion when all 2016 time steps (one week à five minute steps) of a chronic are executed. The second is early termination due to grid failure, which in most cases is equivalent to a blackout. A major cause is the forced disconnection of lines, leading to a cascading failure. It is induced by the Grid2Op rule that a line is disconnected if $\rho_{max,t}$ stays above 100\% for three consecutive time steps. An immediate game over can also be reached if a generator or a load is disconnected and/or a general islanding occurs in the grid. Finally, an episode may fail if the underlying power flow solver does not converge. This sometimes occurs when the grid is far from its original topology. 
\newline
As the evaluation metric of this paper, we use the same \gls{l2rpn} score \cite{marot2021learning} of the 2022 challenge.\footnote{For a detailed description see \url{https://grid2op.readthedocs.io/en/latest/utils.html#grid2op.utils.ScoreL2RPN2022} (last accessed 20/04/2024).} The scoring is based on the survival of a \gls{d_n}, i.e., an agent that does not take any action in the environment. With the \gls{d_n} score set to $0$, we evaluate an agent based on its survival. The agent gets a negative score up to $-100$ if it performs worse than the \gls{d_n}. If outperforms the \gls{d_n}, the score becomes positive, with $80$ corresponding to a completion of all episodes. To achieve the maximum score of $100$, the agent must also optimize the cost of energy loss and the cost of operation. For testing, we were able to obtain the test scenarios of the original WCCI 2022 challenge, thanks to the courtesy of RTE France. 
\subsection{The CurriculumAgent}
\label{ssec:curriculumagent}
\glsreset{ca}
As part of this work, we are retraining and extending the \gls{ca} described in detail in \cite{lehna2023managing}. The \gls{ca} is a \textit{Teacher-Tutor-Junior-Senior} framework first introduced by \cite{binbinchen} in the 2020 \gls{l2rpn} challenge. 
The framework is now shortly described, however for more information we refer to \cite{lehna2023managing}.
\newline
Due to the high number of substation actions, there is a need to reduce the actions set to a suitable subset. For this the \textit{Teacher} runs with a brute-force approach through various scenarios of the environment and simulates with the Grid2Op method \verb|obs.simulate()| the effect of all substation actions. The most frequent actions are then selected by the researcher and combined in an action subset. 
Next to this general approach, the \textit{Teacher} can be modified to search specific, e.g., actions against adversarial agents\cite{binbinchen} or actions that fulfill the N-1 criterion \cite{lehna2023managing}.
\newline
With the action subset, the rule-based \textit{Tutor} is used to generate experiences for imitation learning. In its interaction with the environment, the greedy \textit{Tutor} is only activated if the $\rho_{max,t}$ is above the threshold of $\rho_{tutor}=0.9$, else $a_{DoNothing}$ is returned instead. Within the greedy search, the agent selects the best action based on the lowest $\hat{\rho}_{max,t+1}$ by simulating the actions with \verb|obs.simulate()| . If there is no immediate danger, i.e., $\rho_{max,t}<0.8$, the \textit{Tutor} checks if for a topology reversal, i.e., if a modified substation can be reverted to its original state without a decrease in stability.
Besides reversion, the \textit{Tutor} also checks if it is possible to reconnect any disconnected lines.
\newline
The next component is the \textit{Junior} agent, which is a simple feed-forward network for the \gls{drl} agent that is trained on the experience of the \textit{Tutor} as an imitation learner. The model weights are later used to initialize the \gls{drl} algorithm (warm start). Given that the \textit{Junior} is a supervised learning model, it is further ideal for hyperparameter search and the scaling of the experience. For this paper we used the \gls{bohb} algorithm \cite{falkner2018bohb} for the hyperparameter search. 
\newline
The last component of the \gls{ca} is the \gls{drl} agent called \textit{Senior}. It requires a modified \gls{drl} environment that encapsulates the Grid2Op environment. The agent is trained using \gls{ppo}\cite{schulman2017proximal} and its hyperparameters are chosen by \gls{pbt}\cite{jaderberg2017population}. After the \gls{drl} algorithm reaches convergence, the model is transferred to a \verb|MyAgent| method, which we call for simplicity \gls{senior} with a action threshold of $\rho_{senior}=0.95$. For $\rho_{max,t}>\rho_{senior}$, we take the policies (action probabilities) of the \gls{drl} agent, sort them by probability and iterate with the \verb|obs.simulate()| through the substation actions until the action is below $\rho_{senior}$. This ensures that no illegal actions are taken. The model is further combined with heuristic strategies: automatic line reconnection, topology reversion at $\rho_{max,t}<0.8$ if possible and $a_{DoNothing}$ for every remaining action as long as $\rho_{max,t}<0.95$.  With all these measures, the \gls{ca} shows a strong performance \cite{lehna2023managing} and is therefore a strong benchmark as \gls{senior} in our work. 
\begin{algorithm}[h]
\setstretch{0.95} 
\begin{algorithmic}
\STATE Set $\Psi \gets [\ \ ]$ and $\Theta \gets [\ \ ] $
\FOR{$chronic$ in $env$}
    \STATE $obs_{t=0} \gets$ \verb|env.reset()|
    \STATE $a_{t=0} \gets a_{Do Nothing}$
    \STATE $\psi_{t=0} \gets \psi_{start}$ from $obs_{t=0}$
    \STATE $done \gets \FALSE$
    \WHILE{\NOT $done$}
        \STATE $a_t \gets$ \verb|agent.act()| with $obs_{t}$
        \IF{$a_t ==$ $a_{Do Nothing}$}
            \IF{$\psi_{t}$ in $\Psi$}
                \STATE $\Psi(ID_{\psi_{t}}) +=1 $ increase counter
            \ELSE
                \STATE $ID_{\psi_{t}}\gets$\verb|set_id|$(\psi_{t})$ 
                \STATE $\Psi(ID_{\psi_{t}}) \gets [\psi_{t},1]$  
            \ENDIF
        \ELSE
            \STATE Record $obs_t$, $a_t$, ID in $\Theta$
        \ENDIF
        \STATE $obs_{t+1},done\gets$ \verb|env.step()| with $a_t$
    \ENDWHILE
\ENDFOR
\STATE Sort $\Psi$ based on counter
\RETURN most frequent topologies$\Psi^*_{sub}$
\end{algorithmic}
\caption{Search method for \gls{tt}}
\label{alg:toposearch}
\end{algorithm}
\section{The Topology Approach}
\label{sec:topo}
\subsection{Target Topologies as an alternative to substation actions}
\label{ssec:targettopo}
We pointed out that researchers often consider substation actions only. However, for the WCCI 2022 grid it is well known that that the grid is quite stable in its base topology, i.e., when all substation buses are set to one, as is the case in Figure \ref{fig:grid}. Similarly, \cite{lehna2023managing} have been able to achieve better performance by reverting their agents back to the base topology when the grid is stable. On this basis, we investigate the hypothesis that not only the base topology but also other specific topologies increase the stability of the power grid. 
Consequently, we want to evaluate the impact of these \glsreset{tts}\gls{tts} on the survival of the power system. This can be useful when the base topology is unavailable, e.g., when certain lines are disconnected. 
To find these \gls{tts}, we propose a search algorithm, described in pseudocode in Algorithm \ref{alg:toposearch}, based on counting the number of visits from an \verb|agent| in a given topology. We expect better performance for topologies with a higher count since the agents actively try to reach the topologies with their substation actions.
For the search algorithm, an \verb|agent| interacts with the desired Grid2Op environment $env$, which in our case is the greedy \textit{Tutor} agent. As one can see in Algorithm \ref{alg:toposearch}, we first initialize both an empty topology list $\Psi$ and an empty experience list $\Theta$. Next, we iterate through a certain number of chronics of $env$ to ensure enough diversified topologies. For each chronic of length $T$ with time steps $t=0,\dots,T$ we must first reset the environment and save both the current observation $obs_{t=0}$ and the topology $\psi_{t=0}$ of the state. Afterwards, we iterate through the chronic with our agent and check whether its action is a Do-Nothing action $a_{DoNothing}$ or instead a substation or line action. For the first case we have to check whether the topology $\psi_t$ is already part of our topology list $\psi_t \in \Psi$. If it is, it should already have an ID and a counter in $\Psi$, thus we increment the counter by one. If $\psi_t \notin \Psi$, we call the \verb|set_id|$(\psi_{t})$ method to get a new ID and save the new topology in $\Psi$ with a counter of one. There are three reasons for a $a_{DoNothing}$ action. It appears either at the beginning of the chronic or in case the $\rho_{t}$ falls below the threshold of $\rho_{agent}$ again. A third reason is that the \verb|agent| does not find any legal action, but in this case the environment is expected to reach game over soon. Since the frequency of the last reason is relatively low, it will be sorted out later.

In case the action is not $a_{DoNothing}$, we record the first observation $o_t$ where this is the case, the IDs of the end topologies and the actions that led to the specific topologies in our experience list $\Theta$. However, it is possible that the agent can have multiple steps where the action is not $a_{DoNothing}$. So we have a record of the first observation and all subsequent actions together. While, this recording of $\Theta$ is not necessary for this paper, it allows for future work on training topology \textit{Junior} and \textit{Senior} agents.
If enough chronics are run, we can gather the generated data and sort the topologies of $\Psi$ based on their occurrences. 
Subsequently, by selecting the most frequent $M$ topologies, we are able to assemble our set of \gls{tt} $\Psi^*$ with $\psi^*_m$ and $m=1,\dots,M$.

\subsection{Topology Agent}
\label{ssec:topoagent}
With a suitable subset of \gls{tts} $\Psi^*$, we now extend our \gls{ca} to incorporate the topology approach. The underlying idea is to switch to \gls{tts} when the network starts to become unstable, but before a real emergency occurs. Therefore, we set the threshold of our topology agent to $\rho_{topo}=0.85$ for $\rho_{max,t}$. This ensures that the agent reacts faster to imbalances and steers the network to more stable topologies. However, we still keep our \gls{drl} trained \textit{Senior} component with its usual threshold of $\rho_{senior}=0.95$. Thus, in very critical situations, we first try to resolve the situation with single substation actions before returning to topology optimization. The agent also has the line reconnection component, Do-Nothing actions, and topology reversion when the grid is stable. All these methods are combined in the \verb|agent.act()| method of our \gls{topo}, which can be found as pseudo code in the Algorithm \ref{alg:topoagent}. A new feature within the method is the action buffer $B^{act}$. The buffer allows us to store multiple substation actions and execute them sequentially based on their effect on $\hat{\rho}_{max,t+1}$. Through $B^{act}$ we can reach a \gls{tt} $\psi_m$ in multiple steps without breaking the environment rules. 
Within the method we can see that different components are triggered depending on the $\rho_{max,t}$ of the observation. For $\rho_{max,t}>0.95$ the \textit{Senior} component is activated, for $\rho_{max,t}<0.8$ we test for topology reversion and for the interval of $0.85 <\rho_{max,t}<0.95$ the \gls{tt} component is selected.\footnote{We performed an extensive evaluation for different thresholds on the training environment. Our results showed that the interval between $85\% < \rho_{topo} < 95\%$ gave the best result, even though this creates a gap to the topology reversion.} Note that for all other actions, we select the Do-Nothing action. This is also the case if no suitable substation action or topology could be found. We pass all actions to $B^{act}$ and automatically check for line reconnections.
\begin{algorithm}[!ht]  
\setstretch{0.95} 
\begin{algorithmic}
    \REQUIRE{Observation with $\rho_{max,t}$}
    \ENSURE{No previous action in $B^{act}$}
    \IF{$\rho_{max,t}>0.95$}
        \STATE Run \textit{Senior} for substation action
        \STATE $B^{act} \gets a_{Senior}$
    \ELSIF{$\rho_{max,t}>0.85$}
        \FORALL{$m =  1,\ldots,M $}
            \STATE $\psi_t \gets obs_{t}$  Get current topology
            \STATE $a_{topo,m} \gets \psi_m-\psi_t$ 
            \STATE $\hat{\rho}_{max,m} \gets $ Simulate effect of $a_{topo,m}$
            \IF{Any $\hat{\rho}_{max,m} < 0.85$}
                \STATE $B^{act} \gets a_{topo,m}$
            \ENDIF
        \ENDFOR
    \ELSIF{$\rho_{max,t}<0.8$}
        \STATE Search if topology reversion possible 
        \STATE $B^{act} \gets a_{reversion}$
    \ELSE
        \STATE $B^{act} \gets a_{DoNothing}$
    \ENDIF
    \RETURN $B^{act}$
\end{algorithmic}
\caption{\\ Topology Agents act method}
\label{alg:topoagent}
\end{algorithm}
In the \gls{tt} component, we iterate through the $M$ different \gls{tts} with a simple greedy approach. For each target topology $\psi_m$, we first extract from the current topology $\psi_t$ the combined substation actions required to reach the new topology, denoted as $ a_{topo,m}$. We then take the combined actions and simulate their effect on the grid. However, since \verb|obs.simulate()| only allows one substation change action per time step, we use Grid2Op's Simulator class instead, which supports multiple topology changes at once.\footnote{See \url{https://grid2op.readthedocs.io/en/latest/simulator.html}} 
If we were able to find a topology that satisfies $\hat{\rho}_{max,m}<\rho_{topo}$, we select that topology and pass the combined substation actions to the $B^{act}$. The \gls{topo} can then execute these consecutive steps to reach the new \gls{tt}. With this algorithm set, we are now interested in the effect of \gls{tts}. 
\begin{landscape}
\begin{figure}
    \centering
    \includegraphics[trim={0.0cm 0.0cm 0.0cm 0.0cm},clip,width=25cm]{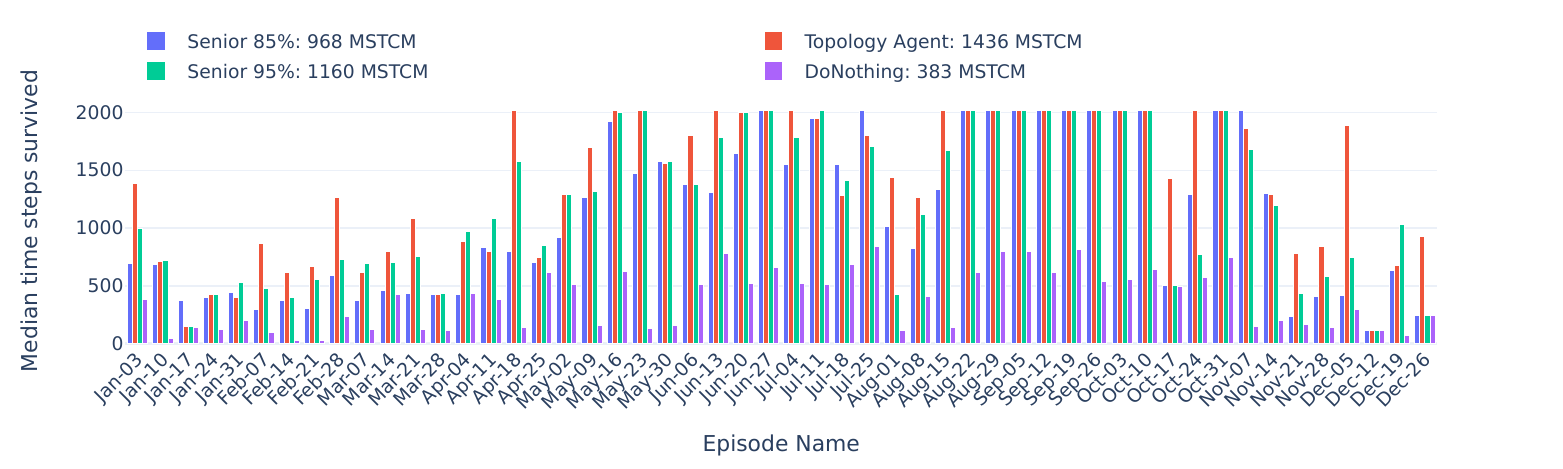}
    \caption{Display of the agent's median survival time across all scenarios of the WCCI 2022 validation environment. The median is computed across the 20 random seeds. On top of the Figure, we display the \gls{mstcm}}
    \label{fig:survival_time}
    \includegraphics[trim={0.5cm 0.5cm 2.0cm 0.2cm},clip,width=7cm]{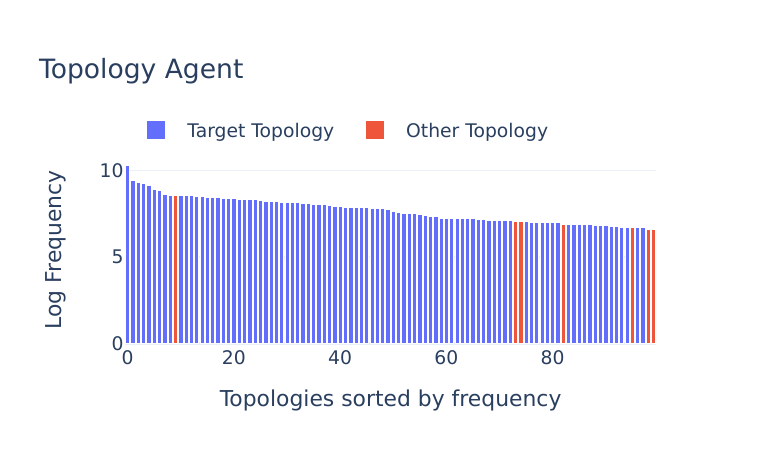}\hfill
    \includegraphics[trim={0.5cm 0.5cm 2.0cm 0.2cm},clip,width=7cm]{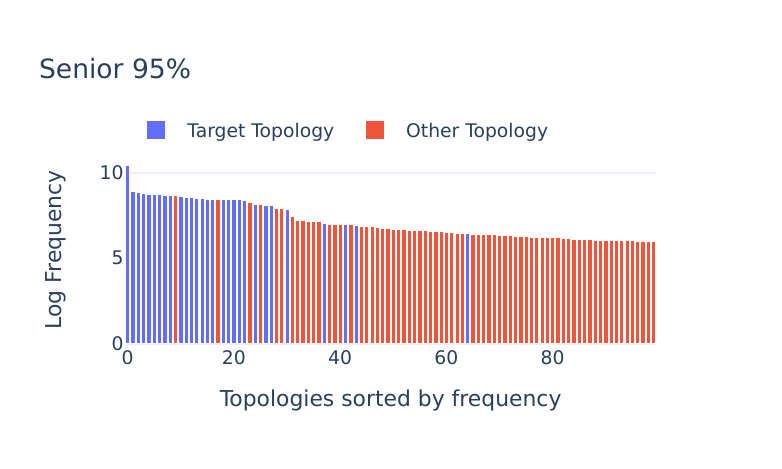}\hfill
    \includegraphics[trim={0.5cm 0.5cm 2.0cm 0.2cm},clip,width=7cm]{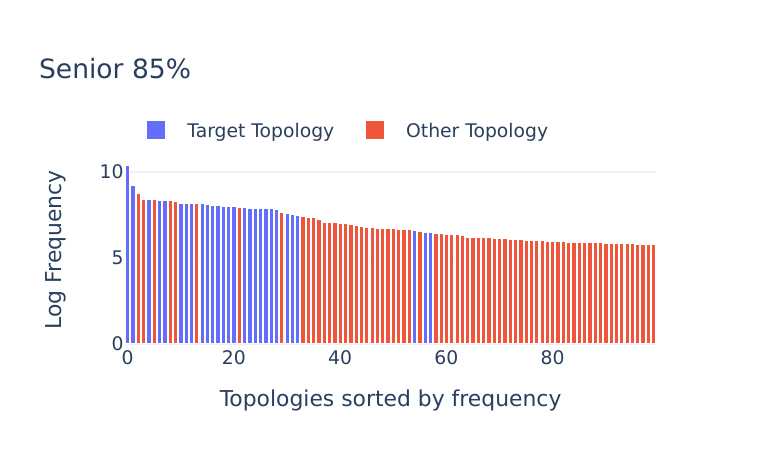}\hfill
    \caption{Frequencies of the most used topologies for all three agents. The topologies are sorted by occurrence on a log-scale. If the topology is a within the set of the 500 \gls{tts}, it is marked in blue otherwise in red. Note that we exclude the base topology for scaling reasons.}
    \label{fig:topologyfreq}
\end{figure} 
\end{landscape}

\section{Experiments}
\label{sec:experiment}
\subsection{Research Design}
\label{ssec:setup}
As described in Section \ref{ssec:grid2op}, we conduct our research on the WCCI 2022 \gls{l2rpn} environment (Figure \ref{fig:grid}), where we train our agents on the publicly available data and used the validation data provided by RTE France. The evaluation set contains a total of 52 scenarios, each consisting of 2016 time steps with a higher \gls{re} mix in comparison to today's power grids. Again, similar to \cite{lehna2023managing}, we propose to use 20 random master seeds to obtain statistically significant results due to the inherent variation between scenarios depending on the environmental seeds.\footnote{The master seeds were randomly chosen by \verb|np.random.choice()| with \verb|np.seed| of $8888$. For each seed with separately copied an validation environment and recomputed its underlying statistics. 
This ensures that our results do not contain any kind of cherry picking.} Regarding the substation action set for our experiments, we did not use the \textit{Teacher} method to generate the actions. Instead, we took the 2000 actions of the 2022 challenge winner \cite{dorfer2022power} and added 30 expert actions selected by RTE, which in total results to an action set of size 2030. 
With this research design, we have chosen a total of four agents to compare the effect of the topology optimization. The advanced agents have the same basic set of substation actions as well as the rule-based components of line reconnection and topology reversion.
\begin{enumerate}
    \item The first agent is the \gls{d_n} baseline, which is expected to score $0.0$.
    \item Second, we propose our previous \gls{senior} from \cite{lehna2023managing} as our benchmark. 
    \item Next we have the new \gls{topo} with the enhancements of Section \ref{ssec:topoagent}. As \gls{tts}, we selected the most frequent 500 topologies that were found with our topology search from Section \ref{ssec:targettopo} on the WCCI 2022 environment with the combined 2030 substation actions. The selection of these 500 \gls{tts} was based on the declining curve of the frequency plot and needs to be adjusted if other environments are selected.
    \item To properly evaluate the effect of the topology approach, we also introduce the \gls{senior85}, which is a retrained \gls{senior} agent with a lower threshold of $\rho_{senior}=0.85$. The reason for this decision is that the \gls{topo} can already start interacting at a $\rho_{max,t} = 0.85$. This way, we enable a fair comparison and make sure that the results of the \gls{topo} are not induced by the lower threshold.
\end{enumerate}
\subsection{Experimental Results}
\label{ssec:results}
\begin{table*}[!ht]
\centering
\begin{tabular}{l|ll|lll|ll}
\toprule
 Agent & Mean & Sd & Median & Q25 & Q75 & \gls{mst} & \gls{mstcm} \\
\midrule
\gls{d_n} & 00.00 & \textbf{0.00} & 00.00 & 00.00 & 00.00 & 229 & 383\\
\gls{senior} & 37.13 &4.49& 37.21 &33.48 & 39.84 & 988 & 1160 \\
\gls{senior85} & 33.07 &3.85 & 32.38 & 30.72 & 36.59 & 806 & 968\\
\gls{topo} & \textbf{41.26} &3.01& \textbf{40.41} &\textbf{39.41} &\textbf{43.69 }& \textbf{1232 }& \textbf{1436}\\
\bottomrule
\end{tabular}
\caption{Summary of the agents' results. All agents are run on the evaluation environment of the WCCI 2022 with 20 different seeds. The performance across the seeds is recorded below. We list the mean and the standard deviation in the first column and the median as well as the 25\% and 75\% quantile in the second column. Further, we depict \glsreset{mst} \gls{mst} and the \glsreset{mstcm}\gls{mstcm} of each agent.}
\label{tab:result_seed_short}
\end{table*}
\paragraph{\gls{l2rpn} Score} With the research design set, we first look at the \gls{l2rpn} score of the 20 seeds that can be found in Table \ref{tab:result_seed_short}. There is a clear increase in the score, as the \gls{topo} agent is able to achieve an average mean score of $41.26$. In comparison, the \gls{senior} reaches a mean score of $37.13$ and the \gls{senior85} only reaches $33.07$. Given that the \gls{topo} agent has the additional \gls{tt} search, it is a clear indication that the topology search might be beneficial. Further, we see that this increase in performance can not be credited to the earlier execution of $\rho_{max,t} = 0.85$, as the \gls{senior85} is performing worse. To ensure significance, we test the $H_0$ hypothesis that the \gls{topo} is from the same distribution as the other agents with the Welch's t-test. The test results are in Table \ref{tab:t_test}, which all reject the $H_0$ hypothesis, indicating significance in the better performance of the \gls{topo}.
This result is not only reflected in the mean, but also by the median values as well. Additionally, looking at the quantiles in Table \ref{tab:result_seed_short}, it becomes clear why multiple seeds are necessary, as a clear variation can be seen between the 25\% and 75\% quantiles. 

\paragraph{Survival Time} Although the score is largely tied to the time steps per chronic, it is quite interesting to look directly at the survival of the agents. For this reason, we added the last two columns of the Table \ref{tab:result_seed_short}. Here, we show the \gls{mst} over all episodes and seeds, where the \gls{topo} is able to reach a median value of $1232$ steps out of $2006$. The \gls{senior} and \gls{senior85} are only able to reach $988$ and $806$ time steps, and the \gls{d_n} agent only survives $229$ steps. As the survival is next to the seeds highly dependent on the chronic, we also conduct the \glsreset{mstcm}\gls{mstcm} in the last column, which is the median survival time after we took the median of the results per chronic. 

As expected, the \gls{mstcm} is higher than the \gls{mst}, since it is not affected as much by outlier performance and instead averages over the chronics first. Nevertheless, we also see a better performance of the \gls{topo} ($1436$ time steps) in comparison to the \gls{senior} $(1160)$, the \gls{senior85} $(986)$ and \gls{d_n} $(383)$.
Next to the two metrics, we also visualize the median performance of the agents for all scenarios of the seeds in Figure \ref{fig:survival_time}. Here, the \gls{topo} shows a much better performance with a total of 17 survived scenarios. In comparison, the \gls{senior} reaches median completion in 12 and the\gls{senior85} in 12 scenarios, while the \gls{d_n} did not survive any chronics. Note that full survival in the median score means that the agent is able to survive at least 50\% of the time. Overall, we argue that optimizing with \gls{tts} increases agent survival on the test data. 

\begin{table}[!hb]
\centering
\begin{tabular}{l|l}
\toprule 
$H_0$ Hypothesis & p-value\\
\midrule 
$H_0: \mu_{DN} = \mu_{Topo_{95}}$ & 2.6e-23 \\
$H_0: \mu_{S_{95}} = \mu_{Topo_{95}}$ & 0.001681\\
$H_0: \mu_{S_{85}}= \mu_{Topo_{95}}$ &7.4e-9\\
\bottomrule
\end{tabular}
\caption{Test Results of the Welch's t-test \cite{welch1947generalization} with the hypothesis $H_0: \mu_i = \mu_j$ against the alternative hypothesis $H_1:\mu_i \neq \mu_j$. For the normality assumption we test with D'Agostino test \cite{d1973tests} and could not reject the $H_0$ hypothesis, so non-normality could not be assumed.}
\label{tab:t_test}
\end{table}
\begin{figure}
    \centering
    \includegraphics[trim={0.6cm 0.8cm 1.8cm 2.5cm},clip, width=1.0\linewidth]{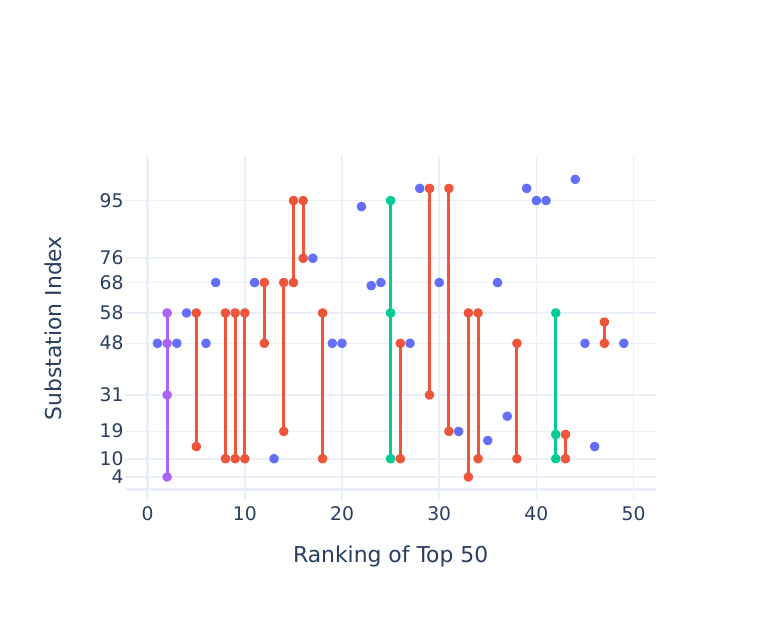}
    \caption{Display of the most frequently used \gls{tts} on the validation data by the \gls{topo}. We rank the topologies based on their occurrence and visualize the effected substations changes. The y-axis shows the switched substation in comparison to the base topology and the x-axis shows the ranked Top50 \gls{tt}.
    The colors indicate the number of changed substations: Blue corresponds to 1, red to 2, green to 3 and purple to 4 changed substation in the \gls{tt}.
    }
    \label{fig:substations}
\end{figure}
\paragraph{Topology Distribution}
Having established that \gls{tts} provide a new strategy for topology optimization, we want to take a closer look at the topologies that were most prominent in the validation set. Therefore, similar to Section \ref{ssec:targettopo}, we counted the number of time steps an agent stays in a topology and sort them by occurrence. The results are visualize in Figure \ref{fig:topologyfreq}, where we show the Top100 topologies and mark in blue if the topology is part of the 500 \gls{tts}. Note that we have excluded the base topology and scaled the y-axis to a logarithmic scale for better visualization. Interestingly, we can observe that \gls{senior} and \gls{senior85} have some topologies from the \gls{tt} set in their Top20 topologies. However, in later topologies there are only marginally some \gls{tts} in their topologies. This may indicate that they select less stable topologies, e.g., topologies that move farther away from the base topology, by using only the substation actions. 
With Figure \ref{fig:substations} we take a closer look at the actual topologies by looking at the Top50 \gls{tts} of the \gls{topo} agent. The y-axis in this Figure denotes the number of substations where the buses have been changed from the base topology. Because different configurations are possible between buses on the same substation, some substations appear more than once, e.g., substation 48. First, we observe that in most cases the \gls{tt} is only one or two changed substations away from the base topology. This supports our understanding that too much deviation from the base topology leads to more imbalance in this specific test setup. Second, we have a few exceptions, with the Top2 \gls{tt} affecting four substations, and the Top25 and Top42 \gls{tts} affecting three substations. In these cases, their stabilizing effect seems to outweigh the deviation from the base topology. A third point to note is that there are some substations that change quite frequently. These are substation 48 with 13 entries in the Top50, followed by substations 58 and 10 with 11 entries, 68 with 8 entries, and 95 with 5 entries. Figure \ref{fig:grid} illustrates the significance of these substations as vital nodes within the electrical grid. 

\begin{figure*}[t]
\begin{subfigure}{.5\textwidth}
  \centering
  \includegraphics[trim={0.3cm 0.0cm  1.2cm 0.0cm},clip, width=1.0\linewidth]{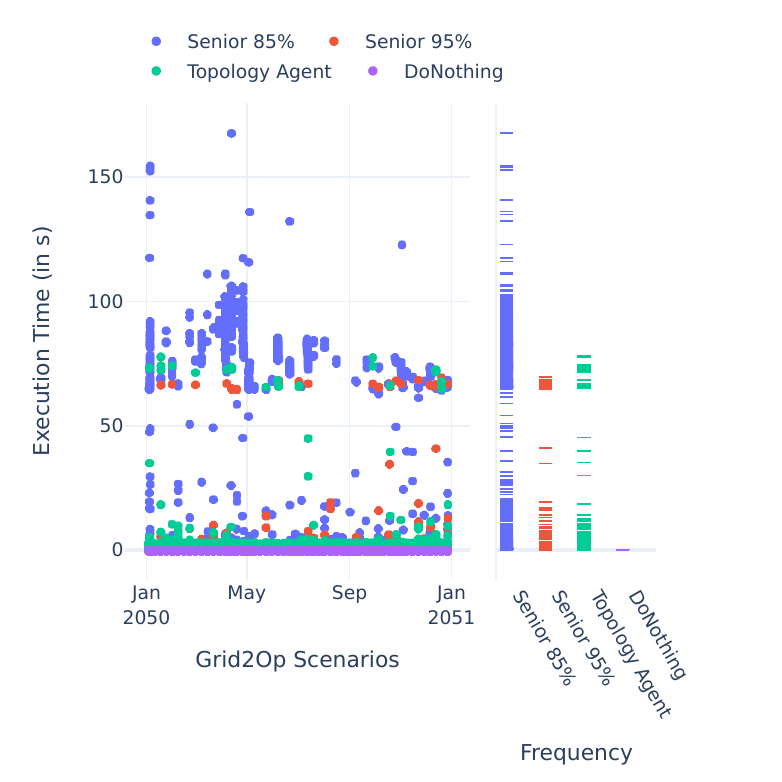}
  \caption{Visualization of the execution time (in s) per action of the four different agents on an exemplary seed run. The left image shows the execution time across all chronics over the year. The right image summarizes the results in a rug plot.}
  \label{fig:executiontime}
\end{subfigure}  \ 
\begin{subfigure}{.5\textwidth}
  \centering
  \includegraphics[trim={0.6cm 0.0cm  1.5cm 0.0cm},clip, width=1.0\linewidth]{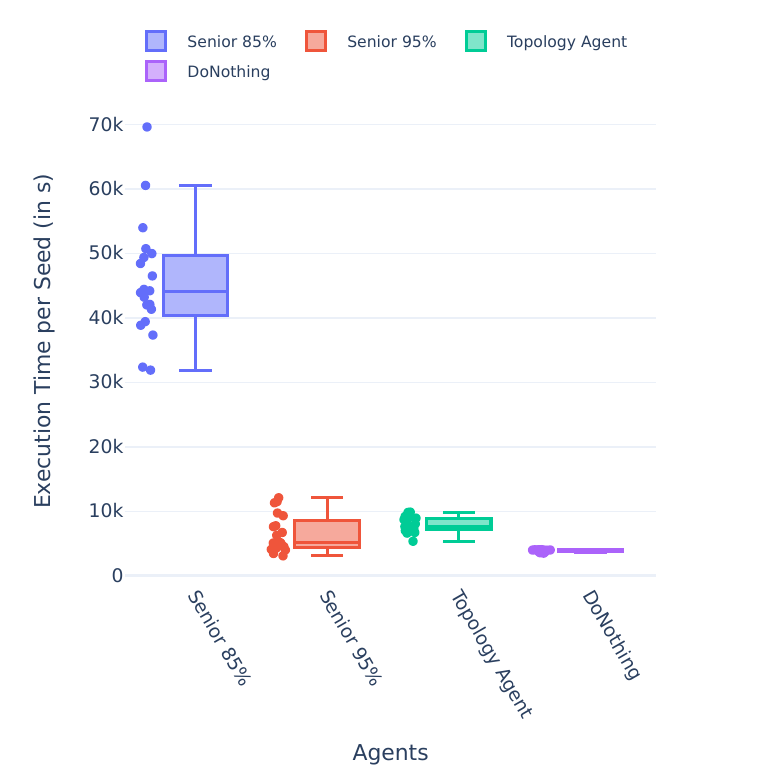}
  \caption{Boxplot of the computation time for each agent. Each dot represents the overall computation time (in s) for one respective seed across all chronics. }
  \label{fig:runtime}
\end{subfigure} \ 
\caption{Execution time for one seed and boxplot of overall computation time across all seeds. }
\label{fig:computation}
\end{figure*}

\paragraph{Computation Time}
Lastly, we want to look at the computation time of the agents. Since we have to iterate through 500 \gls{tts}, we expect a higher computation with the greedy component of \gls{topo}. 
In Figure \ref{fig:executiontime}, we can see the execution time of the four agents on one exemplary seed run. As expected, the \gls{d_n} is the fastest agent, followed by the \gls{senior}. However, in direct comparison, there is only a small increase in execution time for the \gls{topo}. More interestingly, we can see that the \gls{senior85} takes surprisingly a lot longer longer, which could be explained by the more frequent activation due to the lower $\rho_{senior}=0.85$. In addition to the individual results, we can also see that the winter months seem to require longer execution times for the \gls{senior85}, but also partly by the \gls{topo}. Looking at the Figure \ref{fig:survival_time}, we can see that these months are more critical, which explains the longer computation. 
The exemplary result of the one seed is also supported by the box plot of Figure \ref{fig:runtime}. Here we can see that the computation over the seeds is quite large for the \gls{senior85}. Note that while the \gls{topo} generally takes a bit more time, its overall variation in computation time and its maximum value is less than that of the \gls{senior}.

\section{Discussion}
\label{sec:discussion}
After analyzing the experiment results, we were able to show a superior performance of the \gls{topo} agent against the previous \gls{senior} and \gls{senior85}. This is even more remarkable considering that the \gls{senior} with its \gls{drl} agent is already a very strong benchmark for substation actions, line reconnection, and topology reversion.
Nevertheless, both the score and the survival time show a better result of the \gls{topo} over the seeds. Furthermore, as we could see in Figure \ref{fig:survival_time}, we have some scenarios that were solvable in the median survival time by the \gls{topo} agent that the previous agents could not solve. 
This is noteworthy because we did not use a different set of substation actions.
Instead, we used the existing 2030 substation actions to identify suitable \gls{tts} with the novel search approach. Furthermore, the performance of the \gls{senior85} clearly shows that the pure "early" interaction does not correspond to better survival, on the contrary. Therefore, we see the direct topology optimization approach as a possible avenue that should be pursued further. 
Furthermore, we suggest further research in the direction of our topology search. Our experiments indicate that measuring the quality of a topology by the duration an agent stays in it may be beneficial in identifying stable candidates. These topologies seem to create new topology strategies that have not yet been considered, as one could see in Figure \ref{fig:topologyfreq}. In addition, Figure \ref{fig:substations} shows that these strategies are all relatively close to the base topology. This ensures that our desired topologies are still robust. 
With respect to the computation times we expected a higher cost for the greedy iteration. However, it seems based on Figure \ref{fig:executiontime},\ref{fig:runtime} that these costs were only marginal as a more stable grid needed less interaction from the \textit{Senior} component in the \gls{topo} agent. 
Our result on the \gls{tts} correspond to the latest findings of \cite{viebahn2024gridoptions}, who propose with their GridOptions Tool to identify suitable topologies for real world applications. Therefore, the inclusion of \gls{tts} can definitely be recommended. In terms of future work, we already outlined that we save the experience of the topology search. This is of course quite beneficial for future training of \textit{Junior$_{topo}$} and \textit{Senior$_{topo}$} agents as we can use the experience to train the \gls{drl} models. Further, multiple researchers propose a more hierarchical approach \cite{van2023multi,liu2024progressive, manczak2023hierarchical}. This might be interesting when combining them with a topology optimization approach. 

\section{Conclusion}
\label{sec:conclusion}
In this article, we propose a novel addition to the field of topology optimization for power grids. The underlying hypothesis is that certain topologies are more robust than others and should be used for optimization. To this end, we introduce the concept of the \glsreset{tts}\gls{tts} and provide a search algorithm to identify them. We extend our previously developed \gls{ca} to a topology agent \gls{topo}, incorporating the \gls{tts} with a greedy component. The impact of the proposed topology agent on the WCCI 2022 L2RPN environment was analysed in a multi-seed evaluation with 500 \gls{tts}. We found that the \gls{topo} agent outperformed the benchmark by 10\% in \gls{l2rpn} score and by as much as 25\% in median survival time. Further analysis revealed that the \gls{topo} is close to the base topology, which explains the robustness of its performance on the WCCI 2022 environment. Finally, we demonstrate that the incorporation of \gls{tts} as a greedy iteration only marginally increases the execution time. Therefore, we encourage other researchers to pursue the concept of \gls{tts} further, especially in combination with \gls{drl}.

\section*{Acknowledgement}
This work was supported by the research group Reinforcement Learning for Cognitive Energy Systems (RL4CES) from the Intelligent Embedded Systems of the University Kassel and Fraunhofer IEE and the project Graph Neural Networks for Grid Control (GNN4GC) founded by the Federal Ministry for Economic Affairs and Climate Action Germany. 

\section*{Disclosure of Interest}
The authors have no competing interests to declare that are relevant to the content of this article.


 \bibliographystyle{elsarticle-num} 
 \bibliography{hugo_bib}





\end{document}